\documentclass[10pt,twocolumn,letterpaper]{article}

\usepackage[pagenumbers]{cvpr} 

\usepackage{booktabs}
\usepackage{amssymb}
\usepackage[table]{xcolor}
\usepackage{colortbl}
\usepackage{multirow}
\usepackage{graphicx}
\usepackage{xcolor}
\usepackage{multirow}
\usepackage{amsmath}
\usepackage{algorithm}
\usepackage{algorithmic}
\usepackage{makecell}
\usepackage{pgfplots}
\pgfplotsset{compat=1.18}
\usepackage{bbding}
\usepackage{pifont}
\usepackage{float}
\usepackage{threeparttable}
\usepackage{array}
\usepackage{afterpage}

\definecolor{cvprblue}{rgb}{0.21,0.49,0.74}
\usepackage[pagebackref,breaklinks,colorlinks,allcolors=cvprblue]{hyperref}

\def\paperID{310} 
\def\confName{CVPR}
\def\confYear{2026}

\title{ID-Selection: Importance-Diversity Based Visual Token Selection for Efficient LVLM Inference}

\author{
Zhaohong Huang$^{1,*}$, Wenjing Liu$^{1,*}$, Yuxin Zhang$^{1}$, Fei Chao$^{1}$, Rongrong Ji$^{1}$\\[1.0em]
$^{1}$Xiamen University, 361005, P.R. China.
}

\begin{document}
\maketitle
\def\thefootnote{*}\footnotetext{Equal contribution.}
\def\thefootnote{\arabic{footnote}} 
\begin{abstract}
Recent advances have explored visual token pruning to accelerate the inference of large vision-language models (LVLMs).
However, existing methods often struggle to balance token importance and diversity: importance-based methods tend to retain redundant tokens, whereas diversity-based methods may overlook informative ones.
This trade-off becomes especially problematic under high reduction ratios, where preserving only a small subset of visual tokens is critical.
To address this issue, we propose ID-Selection, a simple yet effective token selection strategy for efficient LVLM inference.
The key idea is to couple importance estimation with diversity-aware iterative selection: each token is first assigned an importance score, after which high-scoring tokens are selected one by one while the scores of similar tokens are progressively suppressed.
In this way, ID-Selection preserves informative tokens while reducing redundancy in a unified selection process.
Extensive experiments across 5 LVLM backbones and 16 main benchmarks demonstrate that ID-Selection consistently achieves superior performance and efficiency, especially under extreme pruning ratios.
For example, on LLaVA-1.5-7B, ID-Selection prunes 97.2\% of visual tokens, retaining only 16 tokens, while reducing inference FLOPs by over 97\% and preserving 91.8\% of the original performance, all without additional training.

\end{abstract}   
\section{Introduction}
\label{sec:intro}
In recent years, Large Vision-Language Models (LVLMs) have become a prominent research focus in the multimodal community~\cite{LLaVA,Qwen-vl}.
These models typically consist of a visual encoder \cite{CLIP, SigLip}, a projector, and a large language model (LLM).
The visual encoder extracts image features, which are then mapped by the projector into the semantic space of the LLM~\cite{LLaMA,Qwen} as visual tokens.
These visual tokens are concatenated with instruction tokens and fed into the LLM to generate responses.
Although LVLMs have demonstrated impressive performance across a variety of multimodal tasks~\cite{Gemini, GPT}, they require converting visual inputs into lengthy token sequences.
This significantly increases the inference cost of the model, particularly for high-resolution images \cite{LLaVA-Next, Qwen2.5} and multi-frame videos \cite{Video-LLaVA}.
For instance, Qwen-2.5-VL~\cite{Qwen2.5} handles as many as 16,384 tokens, which is significantly longer than the length of typical text-only sequences.

To address this challenge, recent works~\cite{FastV, FasterVLM, DivPrune, visa} have introduced token pruning strategies to reduce the inference cost of LVLMs by decreasing the number of visual tokens.
\textit{The core of token pruning lies in the selection of tokens, which should yield the highest reduction ratio with the lowest compromise in performance.}
To achieve this, existing methods can be roughly divided into two groups, which we discuss respectively below.

\begin{figure}[t]
    \centering
\includegraphics[width=1\linewidth]{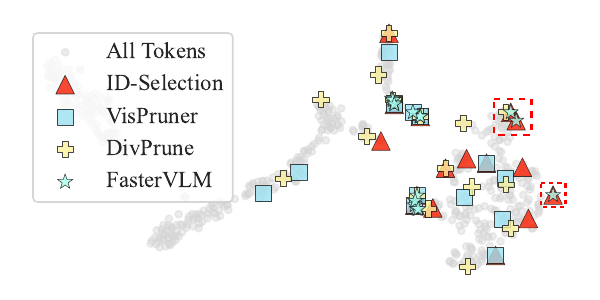}
    \caption{t-SNE visualization of the distribution of 16 selected tokens on POPE~\cite{POPE} using different methods. "All Tokens" represents the original token distribution of LLaVA-1.5-7B~\cite{LLaVA}. Red dashed boxes highlight dispersed tokens that are harder to identify and are often informative for downstream understanding.}
    \label{fig:fig1}
\end{figure}

\textbf{Importance-based Selection:}
These methods typically define an importance criterion for each visual token to quantify its contribution, retaining the most "important" ones while discarding the rest.
For example, FastV~\cite{FastV} assumes that visual tokens with low cross-modal attention scores interact weakly with the instruction and removes them first.
FasterVLM~\cite{FasterVLM}, in contrast, retains tokens with higher \texttt{[CLS]} attention scores, which are assumed to encode more valuable visual cues.
Although these methods can identify informative tokens, they tend to favor tokens from locally dense regions in the feature space.
As shown in Figure~\ref{fig:fig1}, when only 16 tokens are retained, many selected tokens cluster in highly similar regions, resulting in noticeable redundancy, while more dispersed yet informative tokens are often overlooked.

\textbf{Diversity-based Selection:} 
Another line of work aims to eliminate redundant tokens by selecting the most dissimilar ones, thereby ensuring diversity among the retained tokens.
For example, DART~\cite{DART} retains tokens with low duplication based on visual feature similarity.
DivPrune~\cite{DivPrune} iteratively selects tokens that maximize the minimum pairwise distance within the selected subset.
As shown in Figure~\ref{fig:fig1}, although these methods can produce a more dispersed distribution of selected tokens, the retained tokens do not always correspond to the most informative visual content, such as salient visual tokens with high \texttt{[CLS]} attention scores.
As a result, they may preserve visually diverse yet less useful tokens, while overlooking those that are more critical for LLM understanding.


To sum up, for efficient inference in LVLMs, both token importance and diversity are essential to achieve optimal performance under high reduction ratios.
While previous efforts such as VisPruner~\cite{VisPruner} and CDPruner~\cite{CDPruner} have attempted to combine these two factors, they still suffer from different limitations.
VisPruner~\cite{VisPruner} evaluates importance and diversity in a disjoint manner by merging independently selected subsets, and thus fails to jointly balance the two during selection, as shown in Figure~\ref{fig:fig1}.
In contrast, CDPruner~\cite{CDPruner} formulates token selection as a conditional DPP-based~\cite{DPP} subset selection problem.
Although effective, its selection process is relatively more complex, which may limit its practicality in efficient inference scenarios.
Therefore, how to effectively balance token importance and diversity in a simple yet effective manner remains an open challenge for efficient LVLM inference.

To address these limitations, we propose ID-Selection, a simple yet effective token selection strategy that balances importance and diversity for efficient LVLM inference.
The key idea is to couple importance estimation with diversity-aware iterative selection: each visual token is first assigned an importance score, after which high-scoring tokens are selected one by one while the scores of the remaining tokens are updated through a distance-aware suppression function.
Specifically, tokens that are more similar to the selected one receive stronger suppression, whereas those farther away are affected less, enabling the model to progressively reduce redundancy without discarding informative yet distinct tokens.
As shown in Figure~\ref{fig:fig1}, this mechanism helps retain informative tokens in more dispersed regions of the feature space, including those highlighted by the red dashed boxes, which are often overlooked by existing methods.
Since each step only involves a lightweight matrix operation, our method introduces negligible additional inference overhead.
Extensive experiments on \textbf{5} LVLM backbones and \textbf{16} benchmarks demonstrate its strong effectiveness and generality.
For example, on LLaVA-1.5-7B~\cite{LLaVA}, ID-Selection retains only \textbf{16} tokens, reducing visual tokens by \textbf{97.2\%} and inference FLOPs by over \textbf{97\%}, while still preserving \textbf{91.8\%} of the original performance without additional training.
\begin{figure*}[t]
    \centering
    \includegraphics[width=1\linewidth]{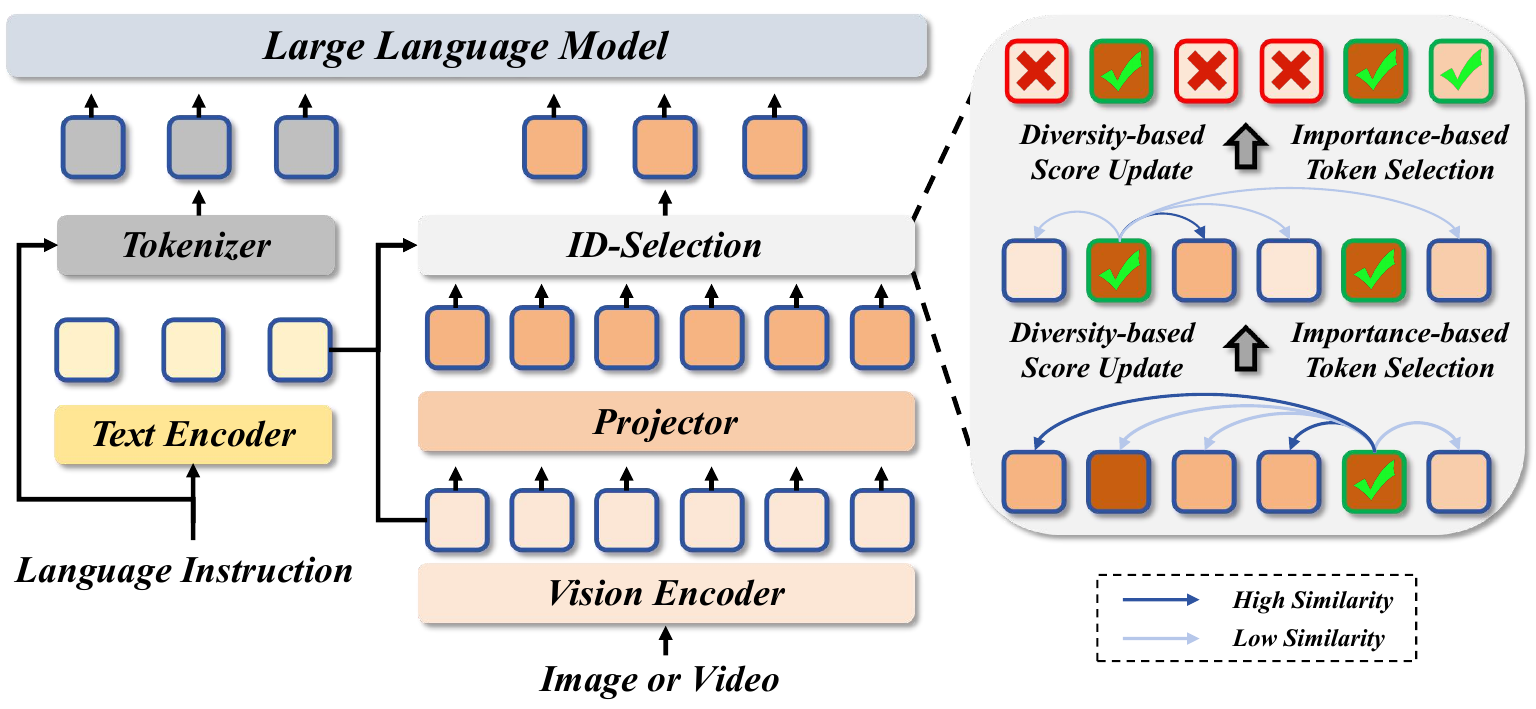}
    \caption{Overview of ID-Selection. Each token is assigned an importance score. For the LLaVA series~\cite{LLaVA,LLaVA-Next,Video-LLaVA}, we adopt a unified score (Eq.~\eqref{eq:6}) that combines visual saliency (\texttt{[CLS]} attention~\cite{FasterVLM}) and instruction relevance (Eq.~\eqref{eq:4}). Tokens with high scores are iteratively selected while suppressing similar ones to maintain diversity, until the token budget is reached, and the remaining tokens are pruned.}
    \label{fig:fig2}
\end{figure*}

\section{Related works}
\label{sec:related_work}

\subsection{Large Vision-Language Models}
Recent advances~\cite{Qwen-vl, Gemini} in Large Vision-Language Models (LVLMs) have significantly improved the integration of visual and linguistic modalities.
For example, LLaVA~\cite{LLaVA} enhances the reasoning ability of large language models on complex tasks such as visual question answering~\cite{sqa, TextVQA} and visual reasoning~\cite{mmbench} by combining the CLIP visual encoder~\cite{CLIP} with the LLaMA language decoder~\cite{LLaMA}. 
However, LVLMs still suffer from hallucination issues~\cite{OPERA}, especially on fine-grained vision-language tasks such as TextVQA~\cite{TextVQA}. 
Recent work~\cite{LLaVA-Next, Feast} attempts to mitigate this by increasing input image resolution to provide finer visual details.
In addition, to further expand the application scope of LVLMs, Video-LLaVA~\cite{Video-LLaVA} introduces video understanding capabilities by processing sequential visual frames.
Yet, both high-resolution images and multi-frame inputs lead to longer visual token sequences, which significantly increase the computational overhead.
Therefore, developing efficient LVLMs has become an increasingly urgent challenge.

\subsection{Visual Token Pruning}
Images often exhibit more redundancy than text~\cite{cfvit}.
As a result, many works have attempted to optimize the inference of LVLMs by reducing the number of visual tokens, with a core focus on token selection.
Existing methods can be broadly categorized into two groups: importance-based selection and diversity-based selection.
Importance-based selection assigns an importance score to each token, retaining those with higher scores while pruning the rest. 
FastV~\cite{FastV} is a pioneering work in this category, pruning tokens with low cross-modal attention after the second layer of the LLM.
Subsequently, some methods~\cite{FasterVLM,VisionZip,Prumerge} propose using \texttt{[CLS]} attention scores as importance metrics.
%
However, these approaches often retain redundant tokens, leading to performance degradation under high reduction ratios.
In contrast, diversity-based methods~\cite{DivPrune,DART} prune visual tokens by maximizing dissimilarity between tokens.
For instance, DivPrune~\cite{DivPrune} formulates the subset selection as a Max-Min Diversity Problem~\cite{MMDP}.
%
%
However, these paradigms may lead to suboptimal performance by overlooking informative yet similar visual tokens.
Recent efforts~\cite{VisPruner,CDPruner} have sought to integrate both factors. 
Nevertheless, they remain limited by suboptimal performance or efficiency during inference.
In this work, we propose ID-Selection, a token selection method that balances importance and diversity.
We first identify informative tokens based on their importance scores and suppress the scores of similar tokens to encourage diversity.
Notably, our method maintains competitive performance under high reduction ratios.
\section{Method}
\label{sec:method}
\subsection{Revisiting Visual Token Pruning of LVLMs}
\label{sec:3.1}
A typical LVLM consists of three components: a visual encoder, a projector, and a Large Language Model (LLM).
Given an input image, the visual encoder, typically a pretrained Transformer~\cite{Transformer} such as CLIP~\cite{CLIP}, transforms it into a sequence of visual patches,
\( \boldsymbol{H}_v = [\boldsymbol{h}_{[cls]}; \boldsymbol{h}_{img}^1, \boldsymbol{h}_{img}^2, \ldots, \boldsymbol{h}_{img}^n] \in \mathbb{R}^{(n+1)\times d} \),
where \(n\) denotes the number of image patches and \(d\) is the output dimension of the visual encoder.
The projector then maps these visual patches into the embedding space of the LLM~\cite{LLaVA,Qwen}, yielding visual tokens \( \tilde{\boldsymbol{H}}_v \in \mathbb{R}^{(n+1)\times D} \), where \(D\) is the input dimension of the LLM.
Finally, the visual tokens are concatenated with instruction tokens and fed into the LLM to generate the response.

While LVLMs excel at multimodal understanding, their inference efficiency is hindered by the large number of visual tokens, which substantially increases computational cost.
To alleviate this burden, recent efforts~\cite{FastV,FasterVLM,DART,DivPrune} have proposed token pruning strategies that improve efficiency by reducing the number of visual tokens.
A common paradigm is to assign each token an importance score and retain only the most important ones.
Such scores are typically derived from either the global attention of the visual encoder or the cross-modal attention within the LLM.

The global attention is usually computed by transforming visual patches \(\boldsymbol{H}_v\) into query \(\boldsymbol{Q}\), key \(\boldsymbol{K}\), and value \(\boldsymbol{V}\) through three weight matrices \(\boldsymbol{W}_Q\), \(\boldsymbol{W}_K\), and \(\boldsymbol{W}_V \in \mathbb{R}^{d \times d}\), respectively:
\begin{equation}
\boldsymbol{Q}=\boldsymbol{H}_v \boldsymbol{W}_Q, \quad \boldsymbol{K}=\boldsymbol{H}_v\boldsymbol{W}_K, \quad \boldsymbol{V}=\boldsymbol{H}_v \boldsymbol{W}_V.
\label{eq:1}
\end{equation}
Then, the scaled dot-product attention is calculated as
\begin{equation}
\boldsymbol{A}=\operatorname{Softmax}\left(\frac{\boldsymbol{Q} \boldsymbol{K}^{\top}}{\sqrt{d}}\right).
\label{eq:2}
\end{equation}
Here, \( \boldsymbol{A} \in \mathbb{R}^{B \times N \times N} \), where \( B \) is the batch size and \( N \) is the visual sequence length.
For the LLaVA family~\cite{LLaVA, LLaVA-Next, Video-LLaVA}, the first row of \(\boldsymbol{A}\) corresponds to the attention from the \texttt{[CLS]} token to all visual tokens, denoted as \(\boldsymbol{S}_{\texttt{[CLS]}} = \boldsymbol{A}[:,0,:]\).
%

In contrast, cross-modal attention~\cite{FastV,SparseVLM} is typically computed between all visual tokens and the last instruction token in the early layers of the LLM.
Given the input sequence \(\boldsymbol{X}=[\boldsymbol{x}_{sys}; \boldsymbol{x}_{img}; \boldsymbol{x}_{str}] \in \mathbb{R}^{l \times D}\), the model transforms it into the query \(\boldsymbol{Q}_{str}=\boldsymbol{x}^{last}_{str}\boldsymbol{W}_{Q}\) and the key \(\boldsymbol{K}_{img}=\boldsymbol{x}_{img}\boldsymbol{W}_{K}\).
The cross-modal attention is then computed as
\begin{equation}
\boldsymbol{S}_{\texttt{cross}}=\operatorname{Softmax}\left(\frac{ \boldsymbol{Q}_{str}\boldsymbol{K}_{img}^T}{\sqrt{d}}\right),
\label{eq:3}
\end{equation}
where \(l\) denotes the length of the input sequence, including the system tokens, visual tokens, and instruction tokens.

Although importance-based methods can effectively identify informative visual tokens, they often retain tokens that are highly similar to each other, leading to substantial redundancy and performance degradation under high pruning ratios.
To address this issue, recent studies~\cite{DART,DivPrune} have shifted toward diversity-based selection, which favors subsets with lower mutual similarity.
However, by emphasizing diversity alone, these methods may overlook informative tokens, such as instruction-relevant objects, and therefore still yield suboptimal performance.

\begin{figure}[t]
    \centering
    \includegraphics[width=1\linewidth]{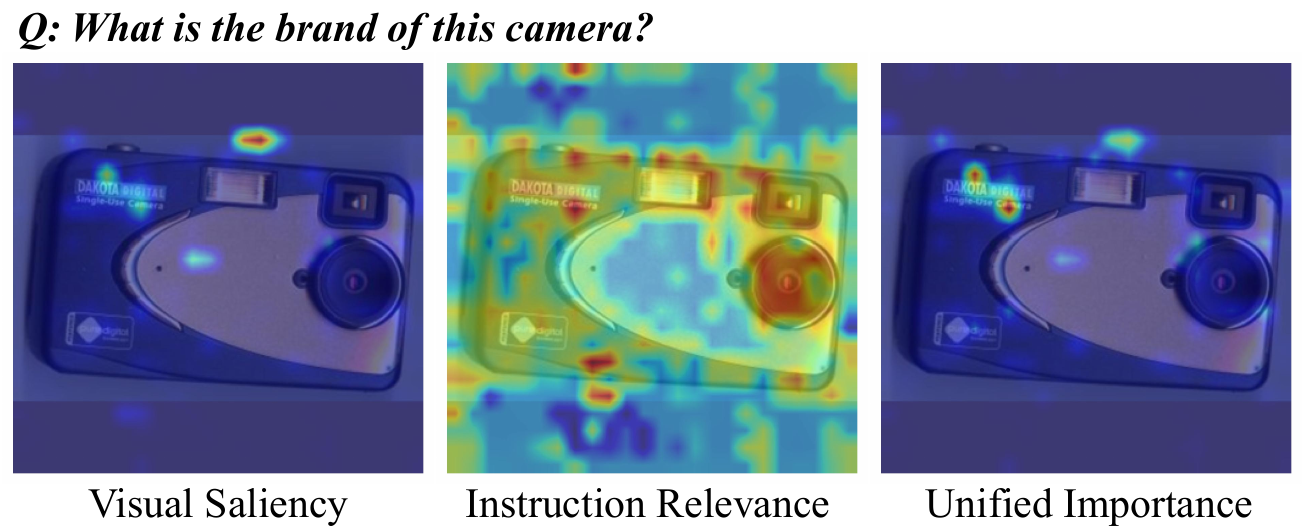}
    \caption{Visualization of visual saliency (\texttt{[CLS]} Attention~\cite{FasterVLM}), instruction relevance (Eq.~\eqref{eq:4}), and unified importance scores (Eq.~\eqref{eq:6}) of LLaVA-1.5-7B~\cite{LLaVA} on TextVQA~\cite{TextVQA}. Red indicates higher scores, while blue indicates lower scores.}
    \label{fig:fig3}
\end{figure}

\subsection{ID-Selection}
In this paper, we introduce ID-Selection, a token selection method that reduces the inference cost of LVLMs by retaining a subset of visual tokens that are both informative and diverse.
The core of ID-Selection is a sequential score-updating process that couples importance estimation with diversity-aware iterative selection.
Specifically, each visual token is first assigned an importance score to identify informative candidates.
The method then iteratively selects the token with the highest current score and updates the scores of the remaining tokens according to their similarity to the selected one, so that redundant tokens are progressively deprioritized during selection.
In this way, ID-Selection preserves informative tokens while reducing redundancy in a unified iterative process.
The overall framework is illustrated in Figure~\ref{fig:fig2}.

\textbf{Importance estimation.} As discussed in Sec.~\ref{sec:3.1}, informative visual tokens are commonly identified using either global attention or cross-modal attention.
The former captures visually salient tokens with rich contextual information during visual encoding, whereas the latter highlights tokens that exhibit strong visual--instruction interactions in the early layers of the LLM.
Accordingly, we instantiate token importance differently depending on the underlying LVLM architecture.

{
\renewcommand{\baselinestretch}{1.2}
\begin{algorithm}[t]
\caption{ID-Selection}
\label{alg:id_selection}
\begin{algorithmic}[1]
\STATE \textbf{Input:} initial scores $\boldsymbol{S}$, visual tokens $\tilde{\boldsymbol{H}}_v$, target token number $T$
\STATE \textbf{Initialize:} selected index set $\mathbf{R} \leftarrow \emptyset$
\WHILE{$|\mathbf{R}| < T$}
    \STATE $i \leftarrow \arg\max_j S_j$
    \STATE $\mathbf{R} \leftarrow \mathbf{R} \cup \{i\}$
    \STATE Compute distances $d(i,j)$ between $\tilde{\boldsymbol{H}}_v^i$ and all remaining tokens $\tilde{\boldsymbol{H}}_v^j$ with $j \neq i$ \hfill (Eq.~\eqref{eq:7})
    \STATE Compute suppression weights $w_{ij}$ \hfill (Eq.~\eqref{eq:8})
    \STATE Update scores for all $j \neq i$: $S_j \leftarrow S_j - w_{ij} \cdot S_i$
    \STATE Set $S_i \leftarrow -\infty$ to prevent reselection
\ENDWHILE
\STATE \textbf{Output:} retain tokens indexed by $\mathbf{R}$ and prune the others
\end{algorithmic}
\end{algorithm}
}

For LLaVA-family models~\cite{LLaVA, LLaVA-Next, Video-LLaVA}, we consider three importance estimators.
The first is the cross-modal attention from the last instruction token in the early LLM layers, following FastV~\cite{FastV}.
The second is the \texttt{[CLS]} attention from the visual encoder, which reflects visual saliency.
The third is a unified importance score that integrates visual saliency and instruction relevance, providing a more holistic criterion for identifying informative tokens.

Specifically, we measure visual saliency using the \texttt{[CLS]} attention \(\boldsymbol{S}_{\texttt{[CLS]}} \in \mathbb{R}^{B \times N}\).
Inspired by~\cite{CDPruner}, we further measure instruction relevance by computing the cosine similarity between the instruction feature \(\boldsymbol{H}_t \in \mathbb{R}^{d}\) and the visual embeddings \(\boldsymbol{H}_v\), formulated as:
\begin{equation}
\boldsymbol{S}_t=\frac{\boldsymbol{H}_v \cdot \boldsymbol{H}_t}{\left\|\boldsymbol{H}_v\right\| \cdot\left\|\boldsymbol{H}_t\right\|}.
\label{eq:4}
\end{equation}
Here, the instruction feature \(\boldsymbol{H}_t\) is extracted from a text encoder paired with the LVLM visual encoder, such as CLIP~\cite{CLIP} or SigLIP~\cite{SigLip}.
We then apply min-max normalization to \(\boldsymbol{S}_t \in \mathbb{R}^{B \times N}\):
\begin{equation}
\tilde{\boldsymbol{S}}_t=\frac{\boldsymbol{S}_t-\min (\boldsymbol{S}_t)}{\max (\boldsymbol{S}_t)-\min (\boldsymbol{S}_t)}.
\label{eq:5}
\end{equation}
The unified importance score \(\boldsymbol{S} \in \mathbb{R}^{B \times N}\) is obtained by element-wise multiplication:
\begin{equation}
\boldsymbol{S}=\tilde{\boldsymbol{S}}_t \times \boldsymbol{S}_{\texttt{[CLS]}}.
\label{eq:6}
\end{equation}
As illustrated in Figure~\ref{fig:fig3}, visual saliency and instruction relevance emphasize different yet complementary regions.
Their combination allows the unified importance score to better highlight regions that are both visually salient and relevant to the input question, thereby providing a more balanced criterion for identifying informative visual tokens.


For architectures without a paired text encoder, such as Qwen2.5-VL~\cite{Qwen2.5} and InternVL2.5~\cite{InternVL2.5}, the unified score in Eq.~\eqref{eq:6} is not applicable.
In such cases, we use the cross-modal attention from the last instruction token as the default importance estimator.


    
    
    
    
    
    

\begin{table*}[t]
\centering
\renewcommand{\arraystretch}{1}
\setlength{\tabcolsep}{2pt}
\caption{Comparison with previous approaches on LLaVA-1.5-7B~\cite{LLaVA} across \(9\) image understanding benchmarks. The highest score is denoted in ~\textbf{bold}. Average are computed by calculating the mean accuracy across all \(9\) datasets. ${\dag}$: importance from LLM second-layer cross-modal attention (Eq.~\eqref{eq:3}); ${\ddag}$: importance from visual encoder \texttt{[CLS]} attention~\cite{CLIP}; ${\S}$: importance from the unified score in Eq.~\eqref{eq:6}.
}
\begin{tabular}{l|c|ccccccccc|c}
\toprule
\textbf{Method} & \textbf{Venue} & \textbf{VQA$^{\text{V2}}$} & \textbf{GQA} & \textbf{VizWiz} & \textbf{SQA$^{\text{IMG}}$} & \textbf{VQA$^{\text{Text}}$} & \textbf{POPE} & \textbf{MME} & \textbf{MMB} & \textbf{MMB$^{\text{CN}}$} & \textbf{Average} \\

\rowcolor{gray!50}
\multicolumn{12}{c}{\textit{Upper Bound, All 576 Tokens (100\%)}} \\

\textcolor{gray}{LLaVA-1.5-7B}~\cite{LLaVA} & \textcolor{gray}{\textit{CVPR'24}} & \textcolor{gray}{78.5} & \textcolor{gray}{62.0} & \textcolor{gray}{50.0} & \textcolor{gray}{69.5} & \textcolor{gray}{58.2} & \textcolor{gray}{85.9} & \textcolor{gray}{1862} & \textcolor{gray}{64.3} & \textcolor{gray}{58.3} &  \textcolor{gray}{100.0\%} \\

\rowcolor{gray!50}
\multicolumn{12}{c}{\textit{Retain 64 Tokens ($\downarrow$ 88.9\%)}} \\
FastV~\cite{FastV} & \textit{ECCV'24} & 55.0 & 46.1 & 50.8 & 51.1 & 47.8 & 48.0 & 1256 & 48.0 & 52.7 & 76.7\% \\
FasterVLM~\cite{FasterVLM} & \textit{ArXiv'24} & 72.6 & 55.0 & 53.2 & 68.9 & 55.3 & 76.8 & 1659 & 61.3 & 55.2 & 94.5\%\\
SparseVLM~\cite{SparseVLM} & \textit{ICML'25} & 68.2 & 52.7 & 50.1 & 62.2 & 51.8 & 75.1 & 1505 & 56.2 & 46.1 & 87.3\% \\
PruMerge~\cite{Prumerge} & \textit{ICCV'25} & 67.4 & 54.9 & 52.9 & 68.6  & 53.0 & 77.4 & 1549 & 59.3 & 51.0 & 91.4\% \\
VisionZip~\cite{VisionZip} & \textit{CVPR'25} & 72.4 & 55.1 & 52.9 & 69.0 & 55.5 & 77.0 & 1690 & 60.1 & 55.4 & 94.5\% \\
DivPrune~\cite{DivPrune} & \textit{CVPR'25} & 74.1 & 57.5 & 53.6 & 68.0 & 54.5 & 85.5 & 1622 & 60.1 & 52.3 & 95.1\% \\
DART~\cite{DART} & \textit{EMNLP'25} & 71.3 & 54.7 & 53.5 & \textbf{69.3} & 54.7 & 73.8 & 1765 &59.5 & 54.0 & 94.0\%\\
VisPruner~\cite{VisPruner} & \textit{ICCV'25} & 72.7 & 55.4 & 53.3 & 69.1 & 55.8 & 80.4 & 1699 &61.3 & 55.1 & 95.4\%\\
CDPruner~\cite{CDPruner} & \textit{NeurIPS'25} & 75.4 & 58.6 & 53.4 & 68.1 & 55.3 & \textbf{87.5} & 1688 & 61.1 & 53.2 & 96.6\%\\
\hline
ID-Selection$^{\dag}$ & \textit{Ours} & \textbf{76.2} & \textbf{59.9} & 53.5 & 68.7 & \textbf{56.5} & 83.6 & \textbf{1783} & 61.8 & 54.2 & \textbf{97.6}\%\\
ID-Selection$^{\ddag}$ & \textit{Ours} &75.2&58.1&\textbf{54.1}&68.0&55.7&85.3&1658&61.0&52.4&96.0\%\\
ID-Selection$^{\S}$ & \textit{Ours} & 75.3 & 58.2 & 53.9 & 68.7 & 56.0 & 86.0 & 1734 & 61.3 & \textbf{55.8} & 97.5\%\\

\rowcolor{gray!50}
\multicolumn{12}{c}{\textit{Retain 32 Tokens ($\downarrow$ 94.4\%)}} \\

FastV~\cite{FastV} & \textit{ECCV'24} & 43.4& 41.5 & 51.7 & 42.6 & 42.5 & 32.5 & 1090 & 37.8 & 33.2 & 63.6\% \\
FasterVLM~\cite{FasterVLM} & \textit{ArXiv'24} & 67.6 & 51.7 & 53.0 & 69.2 & 53.9 & 68.8 & 1594 & 58.3 & 52.6 & 90.5\%\\
SparseVLM~\cite{SparseVLM} & \textit{ICML'25} & 58.6 & 48.3 & 51.9 & 57.3 & 46.1 & 67.9 & 1294 & 51.4 & 40.6 & 79.1\% \\
PruMerge~\cite{Prumerge} & \textit{ICCV'25} & 54.9 & 51.1 & 52.8 & 68.5 & 50.6 & 70.9 & 1503 & 56.8 & 47.0 & 86.2\% \\
VisionZip~\cite{VisionZip} & \textit{CVPR'25} & 67.1 & 51.8 & 52.9 & 68.8 & 53.1 & 68.7 & 1532 & 57.7 & 50.3 & 89.3\%\\
DivPrune~\cite{DivPrune} & \textit{CVPR'25} & 71.2 & 54.9 & 53.3 & 68.6 & 52.9 & 81.5 & 1577 & 57.6 & 49.1 & 92.1\% \\
DART~\cite{DART} & \textit{EMNLP'25} & 67.1 & 52.9 & 52.5 & 69.3 & 52.2 & 69.1 & 1532 & 58.5 & 50.0 & 89.4\%\\
VisPruner~\cite{VisPruner} & \textit{ICCV'25} & 67.7 & 52.2 & 53.0 & 69.2 & 53.9 & 72.7 & 1568 & 58.4 & 52.7 & 91.0\% \\
CDPruner~\cite{CDPruner} & \textit{NeurIPS'25} & 73.6 & 57.0 & 53.1 & 69.5 & 53.2 & \textbf{87.9} & 1657 & 59.6 & 49.6 & 94.7\% \\
\hline
ID-Selection$^{\dag}$ & \textit{Ours} & 72.9 & 55.6 & 53.3 & \textbf{69.7} & \textbf{55.5} & 75.4 & 1691 & \textbf{60.9} & 52.1 & 94.2\%\\
ID-Selection$^{\ddag}$ & \textit{Ours} & 72.8 & 54.9 & 53.5 &68.9&54.8&79.2&1618&60.7&52.3& 93.9\%\\
ID-Selection$^{\S}$ & \textit{Ours} & \textbf{73.9} & \textbf{56.8} & \textbf{53.8} & 68.6 & 55.0 & 84.7 & \textbf{1695} & 60.7 & \textbf{52.8} & \textbf{95.7\%} \\

\rowcolor{gray!50}
\multicolumn{12}{c}{\textit{Retain 16 Tokens ($\downarrow$ 97.2\%)}} \\
FasterVLM~\cite{FasterVLM} & \textit{ArXiv'24} & 61.0 & 47.5 & 50.9 & 67.8 & 51.1 & 55.7 & 1418 & 53.4 & 45.7 & 82.7\% \\
DivPrune~\cite{DivPrune} & \textit{CVPR'25} & 65.4 & 51.4 & 51.5 & 69.1 & 49.8 & 68.8 & 1498 & 54.5 & 44.5 & 86.2\% \\
DART~\cite{DART} & \textit{EMNLP'25} & 61.1 & 48.7 & 51.8 & \textbf{69.5} & 48.3 & 49.1 & 1414 & 50.5 & 41.2 & 80.6\% \\
VisPruner~\cite{VisPruner} & \textit{ICCV'25} & 60.2 & 47.4 & 50.3 & 67.5 & 50.3 & 55.6 & 1434 & 50.3 & 40.8 & 80.8\% \\
CDPruner~\cite{CDPruner} & \textit{NeurIPS'25} & 70.0 & \textbf{55.4} & 52.8 & 68.9 & 50.0 & \textbf{86.4} & 1555 & 57.8 & 43.6 & 90.9\% \\
\hline
ID-Selection$^{\dag}$ & \textit{Ours} & 64.9 & 50.4 & 52.8 & 70.7 & 52.5 & 57.5 & 1451 & 57.2 & 47.1 & 86.2\%\\
ID-Selection$^{\ddag}$  & \textit{Ours} & 68.1 & 52.2 &52.8&68.9&52.6&73.8&1538&58.2&49.8&90.1\%\\
ID-Selection$^{\S}$ & \textit{Ours} & \textbf{70.4} & 53.9 & \textbf{53.3} & 68.1 & \textbf{53.0} & 79.7 & \textbf{1570} & \textbf{58.8} & \textbf{50.0} & \textbf{91.8\%} \\
\bottomrule
\end{tabular}
\label{tab:tab1}
\end{table*}

\begin{table*}[t]
\centering
\renewcommand{\arraystretch}{1}
\setlength{\tabcolsep}{2pt}
\caption{Comparison with previous approaches on LLaVA-Next-7B~\cite{LLaVA-Next} across \(9\) image understanding benchmarks. The highest score is denoted in ~\textbf{bold}. Average are computed
by calculating the mean accuracy across all \(9\) datasets. ${\dag}$: importance from LLM second-layer cross-modal attention (Eq.~\eqref{eq:3}); ${\ddag}$: importance from visual encoder \texttt{[CLS]} attention~\cite{CLIP}; ${\S}$: importance from the unified score in Eq.~\eqref{eq:6}.}
\begin{tabular}{l|c|ccccccccc|c}
\toprule
\textbf{Method} & \textbf{Venue} & \textbf{VQA$^{\text{V2}}$} & \textbf{GQA} & \textbf{VizWiz} & \textbf{SQA$^{\text{IMG}}$} & \textbf{VQA$^{\text{Text}}$} & \textbf{POPE} & \textbf{MME} & \textbf{MMB} & \textbf{MMB$^{\text{CN}}$} & \textbf{Average} \\
\rowcolor{gray!50}
\multicolumn{12}{c}{\textit{Upper Bound, All 2880 Tokens (100\%)}} \\
\textcolor{gray}{LLaVA-Next-7B~\cite{LLaVA-Next}} & \textcolor{gray}{\textit{CVPR'24}} & \textcolor{gray}{81.8} & \textcolor{gray}{64.2} & \textcolor{gray}{57.6} & \textcolor{gray}{70.1} & \textcolor{gray}{61.3} & \textcolor{gray}{86.5} & \textcolor{gray}{1851} & \textcolor{gray}{67.4} & \textcolor{gray}{60.6} &  \textcolor{gray}{100.0\%} \\
\rowcolor{gray!50}

\multicolumn{12}{c}{\textit{Retain 320 Tokens ($\downarrow$ 88.9\%)}} \\

FastV~\cite{FastV} & \textit{ECCV'24} & 72.9 & 55.9 & 53.1 & 62.8 & 55.7 & 71.7 & 1661 & 61.6 & 51.9 & 88.7\% \\

FasterVLM~\cite{FasterVLM} & \textit{ArXiv'24} & 75.8 & 58.5 & 57.0 & 68.3 & 57.9 & 80.4 & 1661 & 63.8 & 55.4 & 93.7\%\\

SparseVLM~\cite{SparseVLM} & \textit{ICML'25} & 74.6 & 57.9 & 54.2 & 67.2 & 56.5 & 76.9 & 1688 & 63.1 & 54.5 & 91.9\% \\

PruMerge~\cite{Prumerge} & \textit{ICCV'25} & 75.3 & 58.8 & \textbf{57.7} & 68.1 & 54.0 & 79.5 & 1710 & 63.0 & 55.6 & 93.2\% \\

VisionZip~\cite{VisionZip} & \textit{CVPR'25} & 76.2 & 58.9 & 56.2 & 67.5 & 58.8 & 82.3 & 1702 & 63.3 & 55.6 & 94.2\%\\

DART~\cite{DART} & \textit{EMNLP'25} & 75.7 & 59.5 & 56.8 & 67.5 & 57.6 & 81.0 & 1744 & 64.2 & 55.7 & 94.3\%\\

DivPrune~\cite{DivPrune} & \textit{CVPR'25} & 77.2 & 61.1 & 55.6 & 67.7 & 56.2 & 84.7 & 1755 & 63.9 & 55.7 & 94.9\%\\

VisPruner~\cite{VisPruner} & \textit{ICCV'25} & 76.4 & 59.4 & 57.5 & 68.1 & \textbf{59.0} & 81.3 & 1781 & 64.1 & 56.0 & 95.2\%\\

CDPruner~\cite{CDPruner} & \textit{NeurIPS'25} & 78.4 & \textbf{61.6} & 55.8 & 67.8 & 57.4 & \textbf{87.2} & 1799 & 65.5 & 55.7 & 96.2\%\\


\hline
ID-Selection$^{\dag}$ & \textit{Ours} &78.0&60.6&56.1&68.1&56.3&85.3&1700&62.5&53.4&94.2\% \\

ID-Selection$^{\ddag}$ & \textit{Ours} & 78.2 & 61.0 & 57.4 & \textbf{68.7} & 58.1 & 85.3 & 1753 & 65.4 & \textbf{57.6} & 96.5\%\\

ID-Selection$^{\S}$ & \textit{Ours} & \textbf{78.7} & 61.2 & 55.4 & 68.3 & 58.4 & 86.4 & \textbf{1817} & \textbf{65.4} & 56.8 & \textbf{96.6\%}\\

\rowcolor{gray!50}
\multicolumn{12}{c}{\textit{Retain 160 Tokens ($\downarrow$ 94.4\%)}} \\
FasterVLM~\cite{FasterVLM} & \textit{ArXiv'24} & 71.7 & 55.3 & 56.6 & 67.2 & 56.9 & 75.8 & 1643 & 60.5 & 52.7 & 90.4\%\\

DivPrune~\cite{DivPrune} & \textit{CVPR'25} & 75.0 & 59.4 & 56.0 & 67.3 & 54.2 & 80.2 & 1703 & 62.9 & 55.1 & 92.8\% \\

DART~\cite{DART} & \textit{EMNLP'25} & 72.5 & 56.8 & 56.7 & 67.8 & 54.9 & 75.3 & 1670 & 62.0 & 53.6 & 91.1\%\\

VisPruner~\cite{VisPruner} & \textit{ICCV'25} & 72.8 & 56.7 & \textbf{57.5} & \textbf{69.0} & 57.0 & 74.9 & 1663 & 61.0 & 52.7 & 91.4\%\\

CDPruner~\cite{CDPruner} & \textit{NeurIPS'25} & 76.7 & 60.8 & 55.2 & 67.5 & 55.4 & 86.8 & 1713 & 64.2 & 53.8 & 94.2\%\\

\hline
ID-Selection$^{\dag}$ & \textit{Ours} &76.2&59.4&55.7&67.2&54.7&82.7&1645&60.8&51.1& 91.8\%\\

ID-Selection$^{\ddag}$ & \textit{Ours} & 76.2 & 59.5 & 57.0 & 67.7 & 57.2 & 83.1 & 1751 & 63.7 & 57.4 & 95.0\%\\

ID-Selection$^{\S}$ & \textit{Ours} & \textbf{76.8} & \textbf{60.2} & 56.2 & 68.0 & \textbf{57.6} & \textbf{85.0} & \textbf{1756} & \textbf{63.7} & \textbf{55.7} & \textbf{95.1\%}\\






\bottomrule
\end{tabular}
\label{tab:tab2}
\end{table*}

\textbf{Diversity-aware iterative selection.} While importance-based selection can identify informative tokens, it often retains highly similar ones, resulting in substantial redundancy.
To alleviate this issue, we propose a diversity-aware iterative selection strategy that preserves informative tokens while progressively suppressing redundant ones.
Specifically, we initialize an empty selection set \(\mathbf{R}\).
At each step, we select the token with the highest current score among the remaining candidates and add it to \(\mathbf{R}\).
We then update the scores of the remaining tokens according to their similarity to the selected one, so that tokens that are more redundant receive lower scores in subsequent steps.

Formally, let the currently selected token be \(\tilde{\boldsymbol{H}}_v^i\) and the remaining unselected tokens be \(\tilde{\boldsymbol{H}}_v^j\), where \(j \neq i\).
We first compute the cosine distance between \(\tilde{\boldsymbol{H}}_v^i\) and each \(\tilde{\boldsymbol{H}}_v^j\):
\begin{equation}
d(i,j)=1-\frac{\tilde{\boldsymbol{H}}_v^i \cdot \tilde{\boldsymbol{H}}_v^j}{\left\|\tilde{\boldsymbol{H}}_v^i\right\| \cdot \left\|\tilde{\boldsymbol{H}}_v^j\right\|}.
\label{eq:7}
\end{equation}
Smaller distances indicate that two tokens are more similar, and thus more likely to be redundant.

To suppress such redundancy in a soft and distance-aware manner, we define the suppression weight as
\begin{equation}
w_{ij}=\exp\left(-\gamma \cdot d(i,j)^2\right),
\label{eq:8}
\end{equation}
where \(\gamma\) controls how quickly the suppression decays with distance.
As a result, tokens that are more similar to the selected one receive stronger suppression, whereas those farther away are affected less.

Finally, we update the score of each remaining token by subtracting a weighted influence from the selected token:
\begin{equation}
S_j \leftarrow S_j - w_{ij}\cdot S_i.
\label{eq:9}
\end{equation}
This process is repeated until the target number of tokens is selected, after which the remaining tokens are pruned.
Notably, each iteration only involves similarity computation and score update over the remaining candidates, resulting in a low per-step cost.
The detailed selection process is summarized in Algorithm~\ref{alg:id_selection}.

\subsection{Complexity Analysis}
At each iteration, ID-Selection selects the token with the highest current score, computes its similarity to the remaining candidates, and updates their scores accordingly.
Assuming the original number of visual tokens is \(N\) and the number of retained tokens is \(T\), each iteration incurs \(O(N)\) cost, yielding an overall complexity of \(O(NT)\).
In practice, since \(T \ll N\) under high pruning ratios, the additional overhead is small.
Moreover, ID-Selection only involves lightweight similarity computation and score update, without requiring matrix inversion or global subset optimization, making it especially suitable for high-compression scenarios.

\section{Experiment}
\label{sec:experiment}

\subsection{Experimental Settings}
\textbf{Datasets and Baselines.}
We conduct extensive experiments on \textbf{16} benchmarks spanning image understanding, fine-grained document understanding, and video question answering.
Specifically, for image understanding, we evaluate on 9 standard benchmarks covering visual question answering~\cite{TextVQA,sqa,GQA,vizwiz,VQAV2}, hallucination evaluation~\cite{POPE}, and visual reasoning~\cite{mmbench,MME}.
For fine-grained visual understanding, we additionally evaluate on 4 text-oriented benchmarks: DocVQA~\cite{DocVQA}, InfoVQA~\cite{InfoVQA}, ChartQA~\cite{ChartQA}, and OCRBench~\cite{OCRBench}.
Finally, we validate our method on 3 video question answering benchmarks: TGIF~\cite{TGIF}, MSVD~\cite{MSVD}, and MSRVTT~\cite{MSRVTT}.
To assess the effectiveness of ID-Selection, we compare it with 9 state-of-the-art token pruning methods, including FastV~\cite{FastV}, FasterVLM~\cite{FasterVLM}, SparseVLM~\cite{SparseVLM}, PruMerge~\cite{Prumerge}, VisionZip~\cite{VisionZip}, DART~\cite{DART}, DivPrune~\cite{DivPrune}, VisPruner~\cite{VisPruner}, and CDPruner~\cite{CDPruner}.

\textbf{Implementation Details.}
ID-Selection is a plug-and-play, training free method that can be seamlessly integrated into existing LVLMs.
We implement it on several widely used open-source LVLMs, including LLaVA-1.5~\cite{LLaVA}, LLaVA-Next~\cite{LLaVA-Next}, Video-LLaVA~\cite{Video-LLaVA}, Qwen2.5-VL~\cite{Qwen2.5}, and InternVL2.5~\cite{InternVL2.5}.
For the LLaVA family, we evaluate three importance estimators: the \texttt{[CLS]} attention from the visual encoder~\cite{CLIP}, the cross-modal attention from the second layer of the LLM (Eq.~\eqref{eq:3}), and the unified importance score defined in Eq.~\eqref{eq:6}.
For Qwen2.5-VL and InternVL2.5, we use the cross-modal attention from the second layer of the LLM (Eq.~\eqref{eq:3}) as the default importance estimator.
The hyperparameter \(\gamma\) in Eq.~\eqref{eq:8} is fixed to 20 across all experiments.

\subsection{Quantitative Evaluation}
\label{sec: 4.2}
\textbf{Results on LLaVA-1.5.}
Table~\ref{tab:tab1} compares ID-Selection with competitive baselines on 9 multimodal benchmarks using LLaVA-1.5-7B~\cite{LLaVA}.
Our method consistently outperforms existing approaches under various pruning ratios.
For example, at an 88.9\% pruning ratio, FastV~\cite{FastV}, which directly performs top-k selection based on cross-modal attention, suffers a 23.3\% performance drop.
In contrast, ID-Selection, built on the same importance cue but with a more effective selection strategy, incurs only a 2.4\% decrease and surpasses the state-of-the-art CDPruner~\cite{CDPruner}.
Notably, even with only 16 visual tokens, ID-Selection with the unified importance metric still retains 91.8\% of the original performance.
These results show that balancing importance and diversity enables ID-Selection to remain robust under extreme pruning.

\begin{table}[t]
\centering
\setlength{\tabcolsep}{0.1pt}
\begin{threeparttable}
\caption{Comparison of Video-LLaVA’s performance~\cite{Video-LLaVA} on 3 video understanding benchmarks~\cite{TGIF,MSVD,MSRVTT}. The highest score is denoted in ~\textbf{bold}. ${\ddag}$: importance from visual encoder \texttt{[CLS]} attention~\cite{CLIP}; ${\S}$: importance from the unified score in Eq.~\eqref{eq:6}.}
\begin{tabular}{l|cc|cc|cc|cc}
\toprule
\multirow{2}{*}{\textbf{Method}} &
\multicolumn{2}{c}{\textbf{TGIF}} &
\multicolumn{2}{c}{\textbf{MSVD}} &
\multicolumn{2}{c}{\textbf{MSRVTT}} &
\multicolumn{2}{c}{\textbf{Average}} \\
\cmidrule(lr){2-3}
\cmidrule(lr){4-5}
\cmidrule(lr){6-7}
\cmidrule(lr){8-9}
 & Acc & Score & Acc & Score & Acc & Score & Acc & Score \\
\midrule
\rowcolor{gray!50}
\multicolumn{9}{c}{\textit{Upper Bound, All 2048 Tokens (100\%)}} \\
\textcolor{gray}{Video-LLaVA}&\textcolor{gray}{19.8} &\textcolor{gray}{2.53} &\textcolor{gray}{70.5} &\textcolor{gray}{3.93} &\textcolor{gray}{57.5} &\textcolor{gray}{3.50} &\textcolor{gray}{100\%} & \textcolor{gray}{100\%} \\
\rowcolor{gray!50}
\multicolumn{9}{c}{\textit{Retain 256 Tokens ($\downarrow$ 87.5\%)}} \\
FastV~\cite{FastV}&15.7 &2.51 &67.3 &3.92 &52.9 &3.38 & 88.9\% & 98.5\% \\
FasterVLM~\cite{FasterVLM} &15.5 & 2.41 &69.0 & 3.92&55.3 &3.42 &90.8\% & 97.6\% \\
DivPrune~\cite{DivPrune}&14.8 &2.40 &\textbf{71.6} &\textbf{3.98} &56.9 &\textbf{3.52} &91.8\% & 98.9\% \\
VisPruner~\cite{VisPruner}&14.3 &2.39 &70.2 &3.95 &56.0 &3.44 &89.7\% & 97.8\% \\
CDPruner~\cite{CDPruner}&15.6 &2.42 &71.1 &3.97 &\textbf{57.7} &\textbf{3.52} &93.3\% & \textbf{99.1\%} \\
\hline
ID-Selection$^{\ddag}$&17.0 &2.46 &70.7 &3.96 &56.2 &3.48 &94.6\% & \textbf{99.1\%} \\
ID-Selection$^{\S}$&\textbf{17.3} &\textbf{2.46} &69.9 &3.94 &57.1 &3.49 &\textbf{95.3\%} & \textbf{99.1\%} \\
\rowcolor{gray!50}
\multicolumn{9}{c}{\textit{Retain 128 Tokens ($\downarrow$ 93.75\%)}} \\

FastV~\cite{FastV}&14.1 &2.38 &64.7 &3.79 &50.9 &3.29 & 83.8\% & 94.8\% \\
FasterVLM~\cite{FasterVLM} &13.8 & 2.36 &68.1 & 3.87&52.0 &3.34 & 85.6\% & 95.7\% \\
DivPrune~\cite{DivPrune}&13.7 &2.38 &\textbf{71.2} &3.96 &57.1 & 3.51&89.8\% & 98.4\% \\
VisPruner~\cite{VisPruner}&13.4 & 2.36 &\textbf{71.2} & \textbf{3.97} &54.7 & 3.45 &87.9\% & 97.6\% \\
CDPruner~\cite{CDPruner}&13.8 & 2.37 &70.8 & 3.94 &\textbf{57.2} & 3.48 &89.9\% & 97.8\%\\
\hline
ID-Selection$^{\ddag}$&\textbf{14.6} &\textbf{2.41} &70.4 &3.94 &56.1 &3.45 &90.4\% & \textbf{98.0\%} \\
ID-Selection$^{\S}$&14.4 &2.38 &70.6 & 3.94&57.0 &\textbf{3.49} &\textbf{90.7\%} & \textbf{98.0\%} \\
\bottomrule
\end{tabular}
\label{tab:tab3}
\end{threeparttable}
\vspace{0.3cm}
\end{table}

\textbf{Results on LLaVA-Next.}
Table~\ref{tab:tab2} further evaluates our method on the high-resolution LLaVA-Next-7B~\cite{LLaVA-Next}.
We report results at pruning ratios of 88.9\% and 94.4\%.
ID-Selection consistently achieves the best overall performance across both settings.
In particular, ID-Selection with \texttt{[CLS]} attention retains 95.0\% of the original performance even when only 5.6\% of the tokens are kept, outperforming both CDPruner~\cite{CDPruner} and VisPruner~\cite{VisPruner}.

\textbf{Results on Video-LLaVA.}
Video-LLaVA~\cite{Video-LLaVA} has demonstrated strong baseline performance in video understanding by extracting 8 frames from each video and converting them into 2048 visual tokens.
However, this token representation is highly redundant. 
As shown in Table~\ref{tab:tab3}, ID-Selection reduces the number of visual tokens to 128 or 256 while still outperforming strong baselines.
These results demonstrate that ID-Selection can effectively preserve informative tokens even in video understanding.
Following previous work~\cite{VisPruner}, we report results on the first 1,000 samples of each benchmark due to commercial API limitations.


\textbf{Results on Qwen2.5-VL.}
Beyond the LLaVA series, we further validate the generality of ID-Selection on Qwen2.5-VL-7B~\cite{Qwen2.5}.
As shown in Table~\ref{tab:tab4}, ID-Selection achieves the best overall performance across all 5 image understanding benchmarks under both token reduction ratios.
At a 77.8\% reduction ratio, it retains \textbf{97.9\%} of the original performance, outperforming FastV and DART by \textbf{4.8\%} and \textbf{3.6\%}, respectively.
When the reduction ratio is further increased to 88.9\%, ID-Selection still preserves \textbf{92.6\%} of the full-model performance, maintaining a clear advantage over both baselines.
We further evaluate Qwen2.5-VL on 4 text-focused benchmarks, where preserving fine-grained textual cues is particularly challenging.
As reported in Table~\ref{tab:tabOCR}, ID-Selection consistently outperforms the competing methods across all datasets and both pruning ratios.
Notably, under the more aggressive 88.9\% reduction ratio, FastV retains only \textbf{53.4\%} of the original performance, whereas ID-Selection still preserves \textbf{70.5\%}, demonstrating a substantially stronger ability to retain dense textual information under extreme compression.

\begin{table}[t]
\centering
\renewcommand{\arraystretch}{1}
\setlength{\tabcolsep}{1pt}
\caption{Comparison of Qwen2.5-VL-7B’s and InternVL2.5-8B’s performance~\cite{Qwen2.5,InternVL2.5} on 5 image understanding benchmarks~\cite{POPE,TextVQA,sqa,mmbench,MME}. The highest score is denoted in ~\textbf{bold}. ${\dag}$: importance from LLM second-layer cross-modal attention (Eq.~\eqref{eq:3}).}
\small 
\begin{tabular}{l|ccccc|c}
\toprule
\textbf{Method} & \textbf{VQA$^{\text{Text}}$} & \textbf{POPE} & \textbf{MMB} & \textbf{SQA$^{\text{IMG}}$} & \textbf{MME} & \textbf{Average} \\
\midrule
\textcolor{gray}{Qwen2.5-VL-7B} &\textcolor{gray}{77.7}&\textcolor{gray}{86.0}&\textcolor{gray}{83.9}&\textcolor{gray}{76.6}&\textcolor{gray}{2330}&\textcolor{gray}{100\%} \\
\rowcolor{gray!50}
\multicolumn{7}{c}{\textit{Token Reduction Ratio = 77.8\%}} \\
FastV~\cite{FastV}&75.2&77.6&80.6&74.0&2000&93.1\%  \\
DART~\cite{DART}&72.0&78.5&79.8&75.2&2193&94.3\%  \\
ID-Selection$^{\dag}$&\textbf{76.0}&\textbf{84.1}&\textbf{82.1}&\textbf{77.1}&\textbf{2217}&\textbf{97.9\%}  \\
\rowcolor{gray!50}
\multicolumn{7}{c}{\textit{Token Reduction Ratio = 88.9\%}} \\
FastV~\cite{FastV}&70.8&68.6&73.8&72.0&1798&86.0\%  \\
DART~\cite{DART}&67.1&73.5&74.1&72.6&\textbf{2009}&88.2\%  \\
ID-Selection$^{\dag}$&\textbf{73.8}&\textbf{78.1}&\textbf{80.0}&\textbf{75.5}&1942&\textbf{92.6\%}  \\
\midrule
\textcolor{gray}{InternVL2.5-8B} &\textcolor{gray}{79.3}&\textcolor{gray}{90.5}&\textcolor{gray}{84.4}&\textcolor{gray}{98.0}&\textcolor{gray}{2324}&\textcolor{gray}{100.0\%} \\
\rowcolor{gray!50}
\multicolumn{7}{c}{\textit{Token Reduction Ratio = 77.8\%}} \\
FastV~\cite{FastV} & 73.5 & 89.6 & 80.5 & \textbf{95.3} & \textbf{2253} & 96.3\% \\
DART~\cite{DART} & 74.2 & 89.2 & 80.0 & 93.8 & 2220 & 95.6\% \\
ID-Selection$^{\dag}$ & \textbf{75.9} & \textbf{90.5} & \textbf{80.9} & 94.4 & 2189 & \textbf{96.4\%} \\
\rowcolor{gray!50}
\multicolumn{7}{c}{\textit{Token Reduction Ratio = 88.9\%}} \\
FastV~\cite{FastV} & 60.7 & 86.8 & 75.8 & 88.8 & 2048 & 88.2\% \\
DART~\cite{DART} & 66.7 & 83.2 & 75.0 & 89.4 & 2002 & 88.5\% \\
ID-Selection$^{\dag}$ & \textbf{70.1} & \textbf{89.5} & \textbf{78.8} & \textbf{92.7} & \textbf{2147} & \textbf{93.5\%} \\
\bottomrule
\end{tabular}
\label{tab:tab4}
\end{table}

\textbf{Results on InternVL2.5.}
We further validate the architectural adaptability of ID-Selection on InternVL2.5-8B~\cite{InternVL2.5}.
As shown in the bottom halves of Table~\ref{tab:tab4} and Table~\ref{tab:tabOCR}, ID-Selection consistently achieves the best overall performance across both general and text-focused benchmarks.
On the 5 general image understanding benchmarks in Table~\ref{tab:tab4}, ID-Selection retains \textbf{96.4\%} and \textbf{93.5\%} of the original performance at token reduction ratios of 77.8\% and 88.9\%, respectively, slightly surpassing the competing methods in both settings.
Its advantage becomes more pronounced on text-focused benchmarks.
As shown in Table~\ref{tab:tabOCR}, under the severe 88.9\% reduction ratio, ID-Selection retains \textbf{60.5\%} of the original performance, compared with only \textbf{44.2\%} for FastV and \textbf{48.0\%} for DART.

These results show that the proposed selection strategy generalizes well beyond the LLaVA family and remains effective even on architectures without a paired text encoder.


\textbf{Efficiency analysis.}
To evaluate the efficiency of ID-Selection, we compare it with prior pruning methods on the high-resolution LLaVA-Next-7B model in terms of theoretical FLOPs, total evaluation time on the POPE benchmark, and KV Cache usage.
All experiments are conducted on a single NVIDIA A800-80GB GPU.
As shown in Table~\ref{tab:tab6}, under an 88.9\% pruning ratio, ID-Selection achieves nearly a \textbf{10$\times$} reduction in FLOPs, a \textbf{2.5$\times$} speed-up in total evaluation time, and over \textbf{89\%} KV Cache reduction compared with the full model.
Moreover, ID-Selection runs faster than CDPruner~\cite{CDPruner} while achieving the same FLOPs and KV Cache savings, indicating that its iterative score-suppression mechanism introduces lower practical overhead.
We also observe that incorporating the text encoder for the unified importance score brings only negligible extra cost, as reflected by the close total evaluation time of ID-Selection$^{\ddag}$ and ID-Selection$^{\S}$.

\begin{table}[t]
\centering
\renewcommand{\arraystretch}{1}
\setlength{\tabcolsep}{6pt}
\caption{Comparison of Qwen2.5-VL-7B’s and InternVL2.5-8B’s performance~\cite{Qwen2.5,InternVL2.5} on 4 text-focused image understanding benchmarks~\cite{ChartQA,OCRBench,DocVQA,InfoVQA}. The highest score is denoted in \textbf{bold}. ${\dag}$: importance from LLM second-layer cross-modal attention (Eq.~\eqref{eq:3}).}
\small
\begin{tabular}{l|cccc|c}
\toprule
\textbf{Method} & \textbf{Chart} & \textbf{OCR} & \textbf{Doc} & \textbf{Info} & \textbf{Average} \\
\midrule
\textcolor{gray}{Qwen2.5-VL-7B} & \textcolor{gray}{84.0} & \textcolor{gray}{83.8} & \textcolor{gray}{95} & \textcolor{gray}{81} & \textcolor{gray}{100.0\%} \\
\rowcolor{gray!50}
\multicolumn{6}{c}{\textit{Token Reduction Ratio = 77.8\%}} \\
FastV~\cite{FastV} & 65.4 & 58.7 & 79 & 54 & 74.4\% \\
DART~\cite{DART} & 53.2 & 58.8 & 60 & 40 & 61.5\% \\
ID-Selection$^{\dag}$ & \textbf{72.4} & \textbf{72.8} & \textbf{86} & \textbf{60} & \textbf{84.4\%} \\
\rowcolor{gray!50}
\multicolumn{6}{c}{\textit{Token Reduction Ratio = 88.9\%}} \\
FastV~\cite{FastV} & 43.0 & 42.8 & 60 & 39 & 53.4\% \\
DART~\cite{DART} & 37.4 & 44.2 & 41 & 31 & 44.7\% \\
ID-Selection$^{\dag}$ & \textbf{60.7} & \textbf{62.0} & \textbf{76} & \textbf{45} & \textbf{70.5\%} \\
\midrule
\textcolor{gray}{InternVL2.5-8B} & \textcolor{gray}{84.1} & \textcolor{gray}{80.3} & \textcolor{gray}{92} & \textcolor{gray}{69} & \textcolor{gray}{100.0\%} \\
\rowcolor{gray!50}
\multicolumn{6}{c}{\textit{Token Reduction Ratio = 77.8\%}} \\
FastV~\cite{FastV} & 65.2 & 56.4 & 59 & 40 & 67.5\% \\
DART~\cite{DART} & 65.0 & 59.9 & 58 & 39 & 67.8\% \\
ID-Selection$^{\dag}$ & \textbf{73.2} & \textbf{60.2} & \textbf{68} & \textbf{44} & \textbf{74.9\%} \\
\rowcolor{gray!50}
\multicolumn{6}{c}{\textit{Token Reduction Ratio = 88.9\%}} \\
FastV~\cite{FastV} & 38.2 & 36.2 & 38 & 31 & 44.2\% \\
DART~\cite{DART} & 42.7 & 44.3 & 38 & 31 & 48.0\% \\
ID-Selection$^{\dag}$ & \textbf{64.2} & \textbf{49} & \textbf{51} & \textbf{34} & \textbf{60.5\%} \\

\bottomrule
\end{tabular}
\vspace{0.2cm}
\label{tab:tabOCR}
\end{table}


\begin{table}[t]
    \centering
    \setlength{\tabcolsep}{2pt}
    \caption{Efficiency comparisons on the POPE benchmark~\cite{POPE} using LLaVA-Next-7B~\cite{LLaVA-Next} under an 88.9\% pruning ratio. We report the theoretical FLOPs, actual runtime, and the KV Cache. ${\ddag}$: importance from visual encoder \texttt{[CLS]} attention~\cite{CLIP}; ${\S}$: importance from the unified score in Eq.~\eqref{eq:6}.}
    \resizebox{\columnwidth}{!}{
        \begin{tabular}{lccccccc}
            \toprule
            Method & FLOPs (T)\(\downarrow\) & Total Time (s)\(\downarrow\) & KV Cache \(\downarrow\)\\
            \midrule
            LLaVA-Next-7B & 20.8 (1.0\(\times\)) & 2294 (1.0\(\times\)) & 1510 MB \\
            FastV & 3.3 (6.3\(\times\)) & 1097 (2.1\(\times\)) & 168 MB \\
            DivPrune & 2.1 (9.9\(\times\)) & 992 (2.3\(\times\)) & 160 MB  \\
            CDPruner & 2.1 (9.9\(\times\)) & 1010 (2.3\(\times\)) & 160 MB \\
            ID-Selection$^{\ddag}$ & 2.1 (9.9\(\times\)) & 916 (2.5\(\times\)) & 160 MB \\
            ID-Selection$^{\S}$ & 2.1 (9.9\(\times\)) & 910 (2.5\(\times\)) & 160 MB \\
            \bottomrule
        \end{tabular}
    }
    \vspace{0.2cm}
    \label{tab:tab6}
\end{table}

\subsection{Ablation Studies}
We conduct ablation studies of ID-Selection on GQA~\cite{GQA} and TextVQA~\cite{TextVQA} using LLaVA-1.5-7B~\cite{LLaVA}, where only 16 visual tokens are retained.

\textbf{Ablation of Iterative Selection.}
We first study the effect of the proposed diversity-aware iterative selection by comparing ID-Selection with top-k selection under the same importance estimator.
Specifically, FastV~\cite{FastV} corresponds to top-k selection based on the last-token cross-modal attention, while FasterVLM~\cite{FasterVLM} corresponds to top-k selection based on the \texttt{[CLS]} attention.
Compared with these baselines, ID-Selection$^{\dag}$ and ID-Selection$^{\ddag}$ further introduce iterative score suppression on top of the same importance estimators.
As shown in Tables~\ref{tab:tab1} and~\ref{tab:tab2}, ID-Selection consistently outperforms FastV and FasterVLM under the same token budgets.
These results indicate that the performance gain comes not only from the importance metric itself, but also from the proposed iterative selection mechanism.

\begin{figure}[t]
    \centering
    \includegraphics[width=1\linewidth]{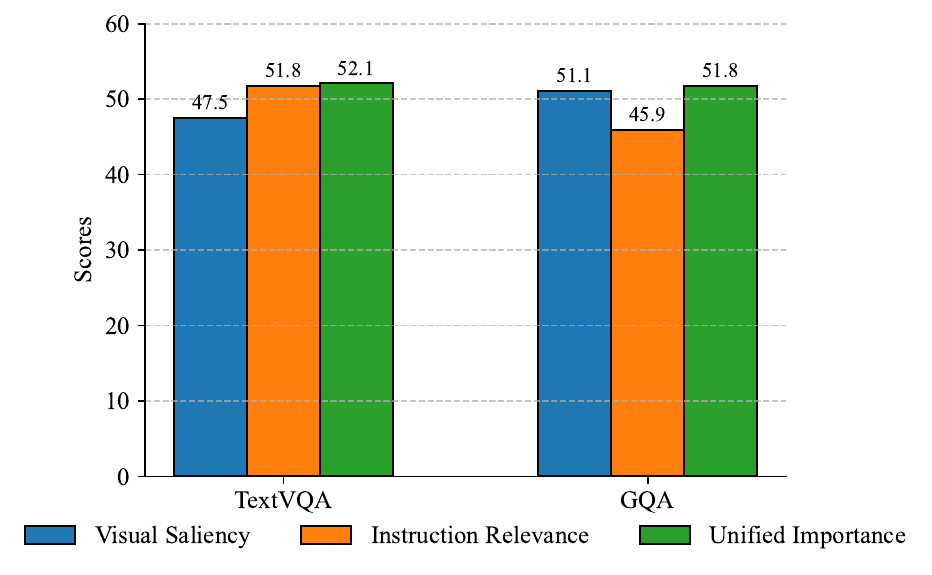}
    \caption{Ablation of Unified Importance.
    }
    \label{fig:fig4}
    \vspace{0.5cm}
\end{figure}

\textbf{Ablation of Unified Importance.}
We then study the effect of different importance estimators by comparing Visual Saliency (\texttt{[CLS]} Attention), Instruction Relevance (last-token cross-modal attention), and Unified Importance (Eq.~\eqref{eq:6}), as shown in Figure~\ref{fig:fig4}.
Visual saliency performs better on TextVQA~\cite{TextVQA}, where fine-grained visual understanding is crucial, whereas instruction relevance is more effective on GQA~\cite{GQA}, where the question provides stronger semantic guidance.
These results indicate that relying solely on visual or textual cues is insufficient.
By integrating both, the unified importance score provides a more balanced criterion for identifying informative tokens across diverse tasks.

\begin{figure}[t]
    \centering
    \includegraphics[width=1\linewidth]{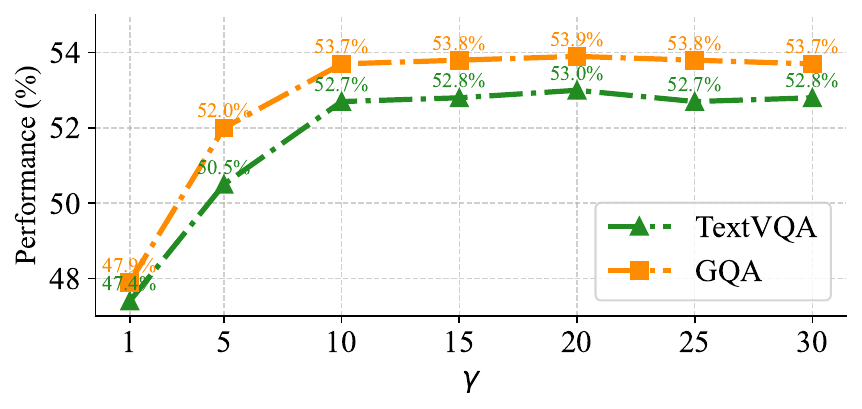}
    \caption{Ablation study on \(\gamma\) in Eq.~\eqref{eq:8}
    }
    \vspace{0.5cm}
    \label{fig:fig5}
\end{figure}

\textbf{Effect of \(\gamma\).}
Finally, we study the parameter \(\gamma\) in Eq.~\eqref{eq:8}, which controls the strength of distance-aware suppression.
As shown in Figure~\ref{fig:fig5}, performance consistently improves as \(\gamma\) increases and gradually saturates afterward.
This suggests that stronger suppression helps remove redundant tokens more effectively, while the saturation behavior indicates that redundancy mainly exists among highly similar tokens.


\section{Conclusion}
In this paper, we propose ID-Selection, a training-free token selection method for efficient LVLM inference.
By coupling importance estimation with diversity-aware iterative selection, ID-Selection preserves informative visual tokens while progressively suppressing redundant ones.
Extensive experiments on \textbf{16} datasets and \textbf{5} LVLMs, covering image understanding, video understanding, and text-rich document understanding, demonstrate that ID-Selection consistently achieves strong performance under various pruning ratios, especially in extreme compression scenarios.
Moreover, ID-Selection significantly improves inference efficiency while introducing negligible additional overhead.
Our results suggest that simple yet effective iterative score suppression offers a practical and general solution for efficient LVLM inference.

{
    \small
    \bibliographystyle{ieeenat_fullname}
    \bibliography{main}
}


\end{document}